# Design and Simulation of an Automated system for quality control in metal canned tuna production using artificial vision

1st Sendey Vera González
Facsistel
Universidad Estatal Península de Santa Elena
La Libertad, Ecuador
svera@upse.edu.ec

2nd Luis Chuquimarca Jiménez
Facsistel
Universidad Estatal Península de Santa Elena
La Libertad, Ecuador
lchuquimarca@upse.edu.ec

3rd Wilson Galdea Gonzalez
Facsistel
Universidad Estatal Península de Santa Elena
La Libertad, Ecuador
wilson.galdeagonzalez@upse.edu.ec

4th Bremnen Véliz
Facsistel
Universidad Estatal Península de Santa Elena
La Libertad, Ecuador
bveliz@upse.edu.ec

5th Carlos Saldaña Enderica
Facsistel
Universidad Estatal Península de Santa Elena
La Libertad, Ecuador
csaldana@upse.edu.ec

*Abstract*— This scientific article presents the implementation of an automated control system for detecting and classifying faults in tuna metal cans using artificial vision. The system utilizes a conveyor belt and a camera for visual recognition triggered by a photoelectric sensor. A robotic arm classifies the metal cans according to their condition. Industry 4.0 integration is achieved through an IoT system using Mosquitto, Node-RED, InfluxDB, and Grafana. The YOLOv5 model is employed to detect faults in the metal can lids and the positioning of the easy-open ring. Training with GPU on Google Colab enables OCR text detection on the labels. The results indicate efficient real-time problem identification, optimization of resources, and delivery of quality products. At the same time, the vision system contributes to autonomy in quality control tasks, freeing operators to perform other functions within the company.

## I. INTRODUCTION

Currently, tuna production companies strive to optimize their processes and reduce production times to achieve their goals [1]. Metal canned goods manufacturers must adhere to strict quality control for export products, where verifying each tuna metal can is essential to maintain better control over each production line [2,3]. The automated system uses a conveyor belt to transport tuna metal cans to the camera [4,5]. Upon detection by the proximity sensor, the image is captured, and at that moment, processing occurs using the learning model through the algorithm programmed in Python on the Raspberry Pi 4 [6,7]. This allows identification and classification based on the type of faults. The learning model enables a review of the hermetic sealing of each metal can and the label text to obtain relevant information about each product [8]. Subsequently, the conveyor belt is activated, moving the metal can to proximity sensor. When the metal can is detected, the classification system is activated, where the robotic arm, using the suction method, extracts the tuna metal cans with the following sequence: those in good condition are placed in the container on the right side, and metal cans with faults such as illegible printed labels, sealing defects on the contours, and unusual positioning of the easy-open lid are placed in the container on the left side.

## II. ARTIFICIAL VISION SYSTEM

### A. Artificial Vision

Artificial Vision is based on images captured by an industrial vision camera and processed through specialized artificial vision software [9]. These systems metal can verify, count, measure, select, and identify faults. Artificial Vision finds applications in various industry sectors, using digital image systems to assess quality control through inspection processes [10,11]. This ensures compliance with minimum quality requirements and reports potential errors in production control. The Artificial Vision System is organized into four pivotal phases. In the first stage, the sensor plays a crucial role in capturing the digital image of the object in question. The second phase focuses on identifying and selecting only the essential parts of the image, thus optimizing the analysis. Subsequently, in the third phase, image processing is conducted using a specialized algorithm, allowing for the extraction of crucial information. Product segmentation is executed precisely during the fourth phase, tailoring the process to the unique characteristics that demand meticulous analysis. This sequential approach ensures an efficient and accurate methodology in applying the Artificial Vision System.

### B. Optical Character Recognition

OCR (Optical Character Recognition) technology utilizes algorithms for image analysis and pattern recognition of characters representing letters, numbers, or other symbols [12]. These characters are compared with an assigned template in a database. OCR systems learn through a neural network, enabling them to recognize characters in various positions, image clarity, or angles [13]. Serialization and traceability are established to achieve production identification, especially in the food industry.

## C. Convolutional Neural Networks

A neural network is a computational model intricately woven with fundamental units known as artificial neurons, typically organized in layers [15]. These interconnected neurons metal can convey a signal comprised of numerical values. The hierarchical structure is delineated into three integral segments: an input layer is responsible for detecting and receiving data, one or more hidden layers for intricate processing, and an output layer that signifies the activation of distinct classes.

## D. YoloV5

This computer vision model is used for object detection, and depending on its version, it offers precision ranging from the most minor (s) to the extra-large (x) [14]. The training time may vary, depending on the version used. The image illustrates the variants of YOLOv5, with its training being faster than EfficientDet. YOLOv5x is the most accurate, capable of quickly processing multiple images.

## III. DEVELOPMENT OF THE PROPOSAL

In this section, the assembly process of the conveyor belt and the robotic arm is detailed, along with their respective sensors, control board, and camera with corresponding connections to implement the artificial vision control system, see Fig 1.

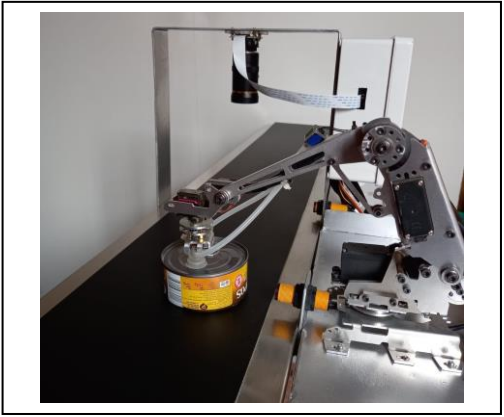

Fig. 1. Artificial Vision Control System

### A. Automatic Conveyor Belt Control

The geared motor is activated for a set time by the microcontroller, which is used to perform the inspection with the vision system. Then, the motor rotates to move the conveyor belt until the metal can's position reaches the classification system. The belt stops for another specified time to allow the robotic arm to select the metal can based on the results of artificial vision.

### B. Photoelectric Sensor Control to Arduino Board

The E18-D80NK sensor is a proximity sensor, NPN type. Internally, it contains a transistor that functions by closing when detection occurs, resulting in a 0V output. However, the output voltage is 5V when no detection occurs.

### C. Servomotor Control with Arduino Board

To initiate the control of the servomotors, we use the 'Servo' library, which simplifies the understanding of actions. Next, we establish the following lines of code to set an initial position for the arm. A specific time is determined for it to execute the following action for the other servo. Servos have a range of 0° to 180°, where it is necessary to define the starting position of the servo and the angle to which it will move within a specified time.

For the development of the artificial vision system, four stages will be employed as detailed below:

- Transport: A conveyor belt will move tuna metal cans to the vision system and then to the classification system.

- Signal Control: An Arduino control board will be used to control the signals and manage the activation of the servo motors in the arm.

- Recognition: An artificial vision system will be implemented using the Raspberry HQ camera and the development of an algorithm for object detection and image processing.

- Classification: The robotic arm performs the classification, determining the location based on established parameters.

### D. Construction of the Dataset for the Artificial Vision Model

We create an images folder within the 'data' file containing all the tuna metal can 'images' we will use. These images were previously captured with the Raspberry HQ camera. Additionally, a folder named 'labels' is created to store annotations. The 'data' folder contains 100% of the images; we must select 80% for training and 20% for validation.

### E. Image Labeling using MAKESENSE Software

- Import Images: Click on "Get Started," which provides the option to import the images for annotating each tuna metal can. Then, click on "Drop images" to proceed.

- Label Creation: Click on "Object Detection," which will open a window for creating labels. Click on the "+" symbol and enter the name for each label.

- Image Labeling: To initiate the Image Labeling project, click on "Start Project" to access the annotation environment tailored for labeling tuna metal cans. "Opt" for the "Rect" option, enabling the delineation of the specific areas of interest within each image. These demarcations identify and classify the various quality states of the tuna product; each assigned its corresponding label. Iterate through this process systematically for every image housed in the designated 'data' folder, ensuring a comprehensive annotation of the entire dataset and facilitating the subsequent analysis of the quality attributes of the labeled tuna metal cans.

- Easy-Open Failure: In this image, we highlight the easy-open feature with a defect in its placement within a rectangle. Its label and outline pass the quality control; see Fig. 2.

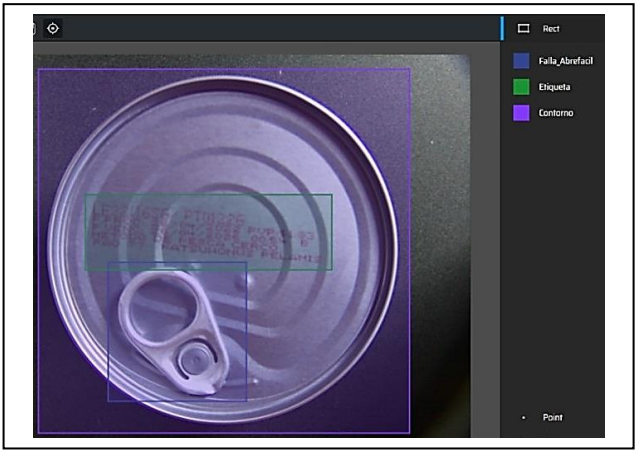

Fig. 2. Easy-Open Failure

- Outline Defect: In this image, we enclose within a rectangle the easy-open feature with a defect in its placement. Its label and outline pass the quality control; see Fig. 3.

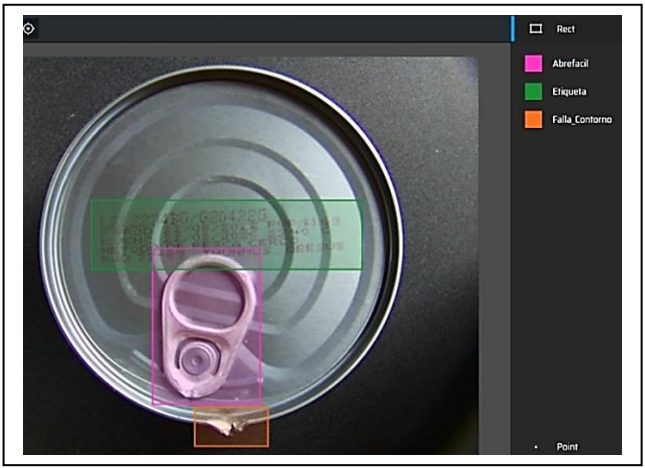

Fig. 3. Outline Defect

- Label Defect: We do not have the label, so we place the rectangle. The easy-open feature and outline pass the quality controls, see Fig. 4.

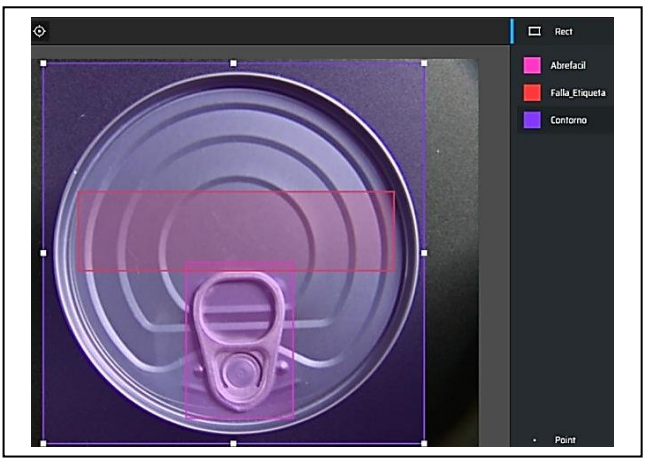

Fig. 4. Label Defect

- Metal can in Good Condition: In the absence of the label, we place the rectangle. The easy-open feature and outline pass the quality controls, see Fig. 5.

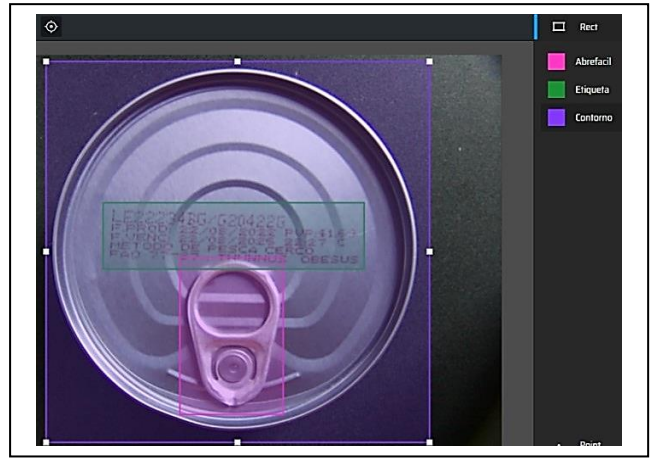

Fig. 5. Metal can in good quality condition

- Exporting Annotations: Once all annotations have been made for each of the images, click on Actions and choose Export Annotations. Select the checkbox for the zip package containing files in YOLO format and click on Export.

### F. Training the Object Detection Model in Google Colab

We begin by accessing the Google Colab platform and adjusting the runtime environment to use the GPU, thus optimizing the program's processing speed. Next, we clone the Ultralytics repository and create the 'yolov5' folder, installing the requirements for training the object detection model. To proceed, we import essential libraries, such as 'torch' and 'utils,' which are crucial in training object detection on images. We use Google Colab to upload the 'data.zip' folder containing images and annotations. We decompress this folder with a single line of code, preparing for the model training. In the newly created 'yolov5' folder, we download and customize the 'coco128.yaml' file. We update the addresses in the 'train' and 'val' sections and input the corresponding labels in the 'names' section. After saving these modifications, we load the adjusted file into Google Colab to initiate the training. We start the training process with everything set up using our own dataset. The relevant line of code runs the 'train.py' algorithm, specifying key parameters such as the image input size, batch size, number of epochs, the configuration file ('custom. yaml' in 'data'), and the location to save the trained models. Finally, upon completing the vision system training in Google Colab, we execute the necessary lines of code to download the 'best.pt' file from the neural network, representing the optimal model obtained during the training process.

### G. Training the Text Recognition Model in Google Colab and Application of EasyOCR

First, we install EasyOCR, which contains all the necessary libraries for model training. Next, we import the EasyOCR library, which is crucial for the model's text recognition. We connect to the EasyOCR folder on Google Drive for file management convenience. Then, we import additional libraries like PIL (Python Imaging Library) and ImageDraw. We display the image to showcase the recognition process. This sequence of steps constitutes the initial setup for text recognition. We perform OCR and obtain bounding boxes for the text located at the top of the label. In the following line of the algorithm, we draw bounding boxes around each recognized line of text on the metal can label, as shown in Fig. 6.

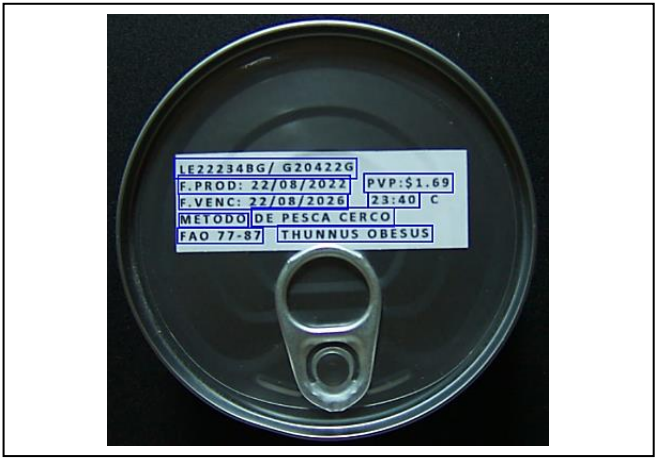

Fig. 6. Text bounding box

Using a for loop with the variable 'i' in the bounds, the recognized text from the trained model is printed.

### H. Classification System with Robotic Arm

For the programming of the sequences performed by the robotic arm, the Arduino IDE software was used, and to make the connections, the Sensor Shield V5.0 board was employed, which provides a better organization of the connections to be made. The vision system, through detection, provides quality data for each metal can. With the classification system of the robotic arm previously programmed in Arduino, it receives the information to control the robot's servo motors. The classification system features a sensor shield board and uses a 5V, 5A voltage source, which is recommended to ensure optimal operation of the robotic arm system. The generation of movements is achieved using servo motors. For this project, we employed MG996R servos to execute the required movements of the robot's base, forearm, and arm. Additionally, MG90S servos were used to control the movements of the wrist and hand of the robotic arm.

### I. Secuencias de funcionamiento del brazo robotico

Each servo motor's position is controlled by angles, with a total rotation of 180°. In the project, we use 90° rotations at each position. The classification system of the robotic arm initiates the sequence from the center of its base with an initial position set at 90° for each of the six servo motors. When the photoelectric sensor located at the bottom of the robotic arm detects a metal can, it initiates the sequence based on the detection performed by the vision system. The robotic arm suctions the metal can if the detection indicates faults in easy-open, contour, and label. It employs the displacement sequence towards its proper end, depositing the product in that container. When the vision system recognizes a metal can meeting quality standards, such as its easy-open feature, label, and contour, the robotic arm sequence is initiated upon the metal can passing through the photoelectric sensor. It begins with a tilt to suction the metal can, followed by an upward movement to rotate 90° to the left. Once in that position, it lowers to release the metal can, then ascends to return to its initial position at the center of its base, set at 90°.

## IV. RESULTS

### A. Testing the algorithm's functionality on the Raspberry Pi 4

In Fig. 7, three detections are observed: a flaw, indicating the absence of a label with a 69% accuracy, a labeled metal can displaying production data and identified as good quality, and an easy-open feature with percentages exceeding 70% and 96%, respectively.

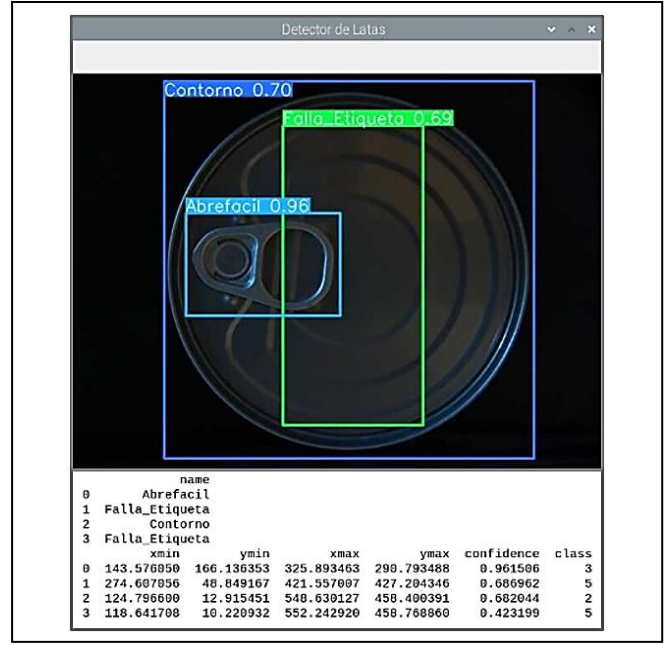

Fig. 7. Inspection Result: Labeling Failure

In Fig. 8, the detection of a flaw with filaments in the sealed contour of the metal can is observed, with a precision percentage of 95%. The label and easy-open feature are also detected, along with the corresponding confidential percentage and bounding box coordinates for each.

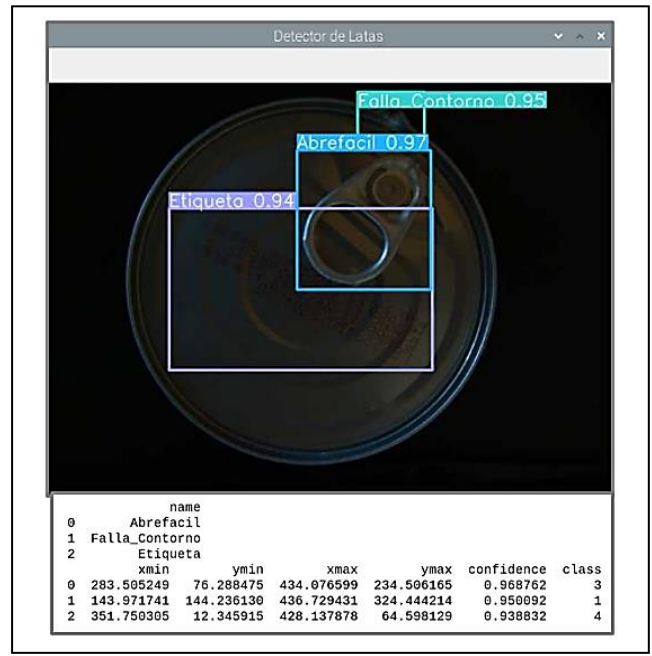

Fig. 8. Inspection Result: Sealing Contour Flaw

In Fig. 9, a flaw in the easy-open feature is detected with a percentage of 96%. When displaced, the detected position could lead to the lid seal opening under pressure. The sealing contour is identified with a confidentiality percentage of 96%, and the label is recognized with 94%. These high percentage values indicate a high level of accuracy in the metal can detections.

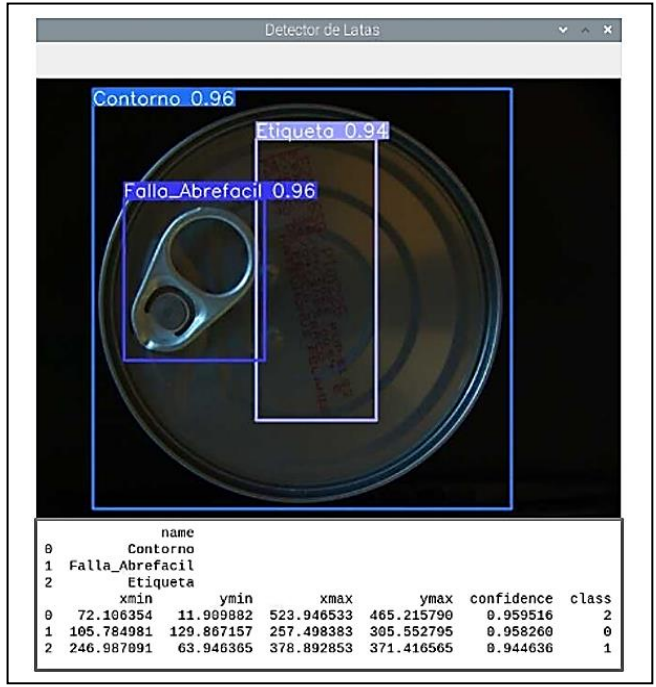

Fig. 9. Inspection Result: Easy-Open Flaw

In Fig. 10, the inspection result shows a metal can with its easy-open feature, contour, and label in optimal quality. At the bottom, informative details such as class, accuracy, and minimum and maximum values for each label are provided.

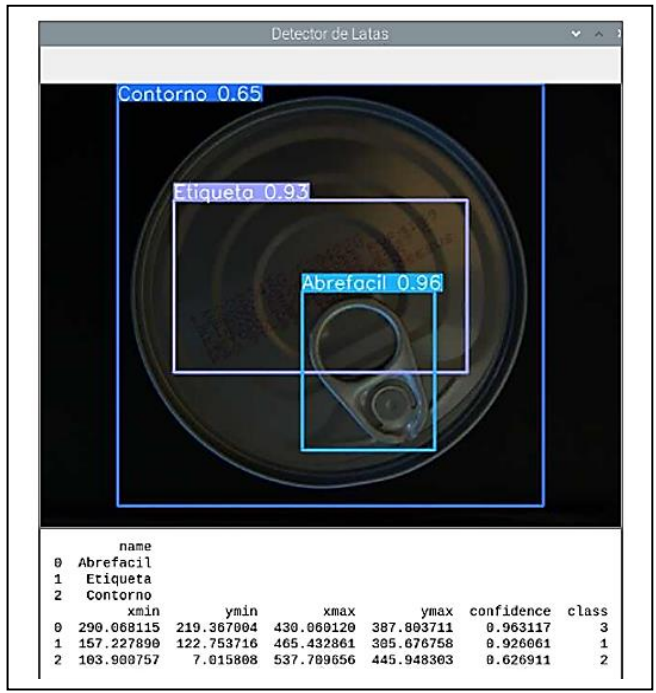

Fig. 10. The inspection result is a metal can in good quality condition

The training results with 200 iterations provide graphs for accuracy, recall, mAP0.5, and mAP0.5:0.95 metrics. These values ensure the model achieves better mean average precision (mAP) values. In the precision metrics, when reaching 200 epochs, results show an accuracy approaching 99%, meaning the bounding boxes were correct when a metal can was detected. In Fig. 11, in the recall metrics (sensitivity metrics), it is observed that, after completing all iterations, the results provide true positives and false negatives. True positives (TP) indicate correct detection in an image, while false negatives (FN) are used to avoid labeling a changed image.

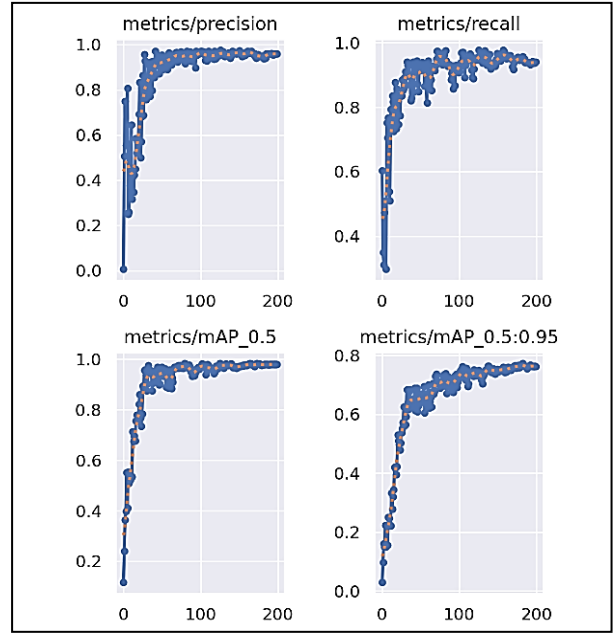

Fig. 11. Graphs of metric/precision, recall/mAP0.5, and mAP0.5:0.95

The mAP0.5 metrics determine values above 0.5 in their detections, considering predictions within an acceptance threshold regarding the detected predictions. On the other hand, the mAP0.5:0.95 metrics, when detecting predictions above 0.95, are considered perfect in their detections.

In the following Fig. 12, we visualize confidence values on the x-axis and model precision data on the y-axis. The blue line represents the total of all classes with a confidence value ranging from 1 down to 0.985 precision. It illustrates how the precision values for each label change as the iterations progress.

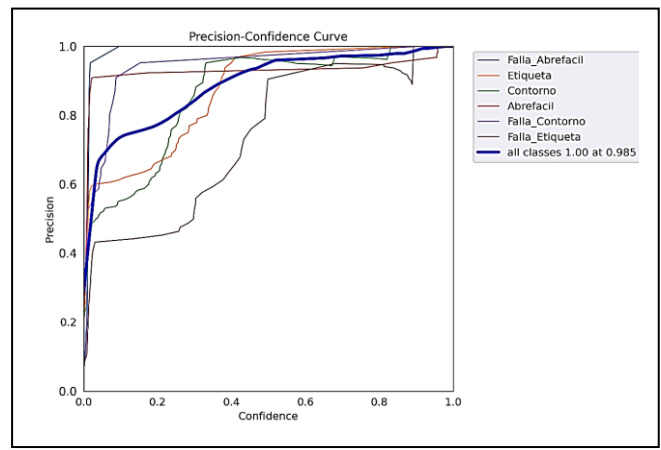

Fig. 12. Confidence and Precision Curve Graph

In training, it is possible to verify metric values such as metrics/precision, which helps determine if the model that performed the metal can detection was accurate. Table 1 details metric data, considering epochs 100 to 199, where a precision percentage of 0.95917 is obtained, indicating a perfect detection rate. The mAP (mean Average Precision) metric chooses the best weights and then calculates on test images. This metric is crucial as it defines if the model is robust enough when performing detections in the vision system.

TABLE I.   TRAINING METRICS IN GOOGLE COLAB WITH 200 EPOCHS

| Epoch | Metrics/ precisión | Metrics/ recall | Metrics/ mAP0.5 | Metrics/ mAP0.5: 0.95 |
|---|---|---|---|---|
| 190 | 0.95869 | 0.93715 | 0.98006 | 0.76361 |
| 191 | 0.95992 | 0.93761 | 0.98025 | 0.76555 |
| 192 | 0.96057 | 0.9319 | 0.98032 | 0.76698 |
| 193 | 0.95859 | 0.93525 | 0.97954 | 0.76375 |
| 194 | 0.95499 | 0.94051 | 0.97913 | 0.76153 |
| 195 | 0.95687 | 0.94146 | 0.97906 | 0.76425 |
| 196 | 0.95982 | 0.93792 | 0.97891 | 0.76249 |
| 197 | 0.95965 | 0.93861 | 0.97891 | 0.76291 |
| 198 | 0.95935 | 0.93749 | 0.97921 | 0.76414 |
| 199 | 0.95917 | 0.94134 | 0.979 | 0.76313 |

For metrics/mAP0.5, the last iteration is taken as a reference, resulting in a precision percentage of 0.979 with high precision and confidence above 0.5. In metrics/mAP0.5:0.95, at epoch 199, a confidence value of 0.95 and a precision value of 0.76313 in object detection are obtained. These values improve when more epochs are added to the training, and losses metal can be reduced.

## V. CONCLUSIONS

In developing this project, the YOLOv5x model was employed for detecting defects at the edges of tuna metal can lids. A dataset was created, and during training, images were identified with rectangles to locate the defects. The expected detection results were achieved through the convolutions performed by this algorithm.

It is necessary to use hardware with significant computational resources to acquire label data from metal cans for OCR text recognition. The training was conducted in Google Colab, which provides a GPU for fast processing iterations. This enabled text recognition and data extraction from labels, applying OCR text detection accordingly.

For real-time detection of the positioning of the easy-open ring, the YOLOv5x model was utilized because it contains a compressed model that allows for increased speed in defect detection according to established standards for metal canned products. Through training in Google Colab, the coordinates for positioning the easy-open ring were determined with a bounding box.

REFERENCES

[1] R. Sunoko and H.-W. Huang, 'Indonesia tuna fisheries development and future strategy', Marine Policy, vol. 43, pp. 174–183, 2014.

[2] M. Doddema, G. Spaargaren, B. Wiryawan, and S. R. Bush, 'Responses of Indonesian tuna processing companies to enhanced public and private traceability', Marine Policy, vol. 119, p. 104100, 2020.

[3] E. R. Blickem, J. W. Bell, D. M. Baumgartel, and J. Debeer, 'Review and Analysis of Tuna Recalls in the United States, 2002 through 2020', Journal of Food Protection, vol. 85, no. 1, pp. 60–72, 2022.

[4] X. Lekunberri, J. Ruiz, I. Quincoces, F. Dornaika, I. Arganda-Carreras, and J. A. Fernandes, 'Identification and measurement of tropical tuna species in purse seiner catches using computer vision and deep learning', Ecological Informatics, vol. 67, p. 101495, 2022.

[5] I. Konukseven, B. Kaftanoglu, and T. Balkan, 'Multisensor controlled robotic tracking and automatic pick and place', in Proceedings of the 1997 IEEE/RSJ International Conference on Intelligent Robot and Systems. Innovative Robotics for Real-World Applications. IROS'97, 1997, vol. 3, pp. 1356–1362.

[6] S. Mathur, B. Subramanian, S. Jain, K. Choudhary, and D. R. Prabha, 'Human detector and counter using raspberry Pi microcontroller', in 2017 Innovations in Power and Advanced Computing Technologies (i-PACT), 2017, pp. 1–7.

[7] H. Meddeb, Z. Abdellaoui, and F. Houaidi, 'Development of surveillance robot based on face recognition using Raspberry-PI and IOT', Microprocessors and Microsystems, vol. 96, p. 104728, 2023.

[8] I. Ilhan, D. Turan, I. Gibson, and R. ten Klooster, 'Understanding the factors affecting the seal integrity in heat sealed flexible food packages: A review', Packaging technology and science, vol. 34, no. 6, pp. 321–337, 2021.

[9] J. Yang, C. Wang, B. Jiang, H. Song, and Q. Meng, 'Visual perception enabled industry intelligence: state of the art, challenges and prospects', IEEE Transactions on Industrial Informatics, vol. 17, no. 3, pp. 2204–2219, 2020.

[10] M. Abd Al Rahman and A. Mousavi, 'A review and analysis of automatic optical inspection and quality monitoring methods in electronics industry', Ieee Access, vol. 8, pp. 183192–183271, 2020.

[11] A. Kazemian, X. Yuan, O. Davtalab, and B. Khoshnevis, 'Computer vision for real-time extrusion quality monitoring and control in robotic construction', Automation in Construction, vol. 101, pp. 92–98, 2019.

[12] A. Chaudhuri et al., Optical character recognition systems. Springer, 2017.

[13] J. Singh and B. Bhushan, 'Real time Indian license plate detection using deep neural networks and optical character recognition using LSTM tesseract', in 2019 international conference on computing, communication, and intelligent systems (ICCCIS), 2019, pp. 347–352.

[14] A. B. Amjoud and M. Amrouch, 'Object Detection Using Deep Learning, CNNs and Vision Transformers: A Review', IEEE Access, 2023.

[15] M. R. G. Meireles, P. E. M. Almeida, and M. G. Simões, 'A comprehensive review for industrial applicability of artificial neural networks', IEEE transactions on industrial electronics, vol. 50, no. 3, pp. 585–601, 2003.